\documentclass[conference]{IEEEtran}
\IEEEoverridecommandlockouts
\usepackage{cite}
\usepackage{amsmath,amssymb,amsfonts}
\usepackage{algorithmic}
\usepackage{graphicx}
\usepackage{textcomp}
\usepackage{xcolor}

\usepackage{nameref}

\usepackage[utf8]{inputenc}
\usepackage{float}
\usepackage{subfig}

\def\BibTeX{{\rm B\kern-.05em{\sc i\kern-.025em b}\kern-.08em
    T\kern-.1667em\lower.7ex\hbox{E}\kern-.125emX}}

\makeatletter
    
\begin{document}

%

\title{Improve High Level Classification with a More Sensitive metric and Optimization approach for Complex Network Building}

\author{
\IEEEauthorblockN{Josimar Edinson Chire Saire}
\IEEEauthorblockA{\textit{Institute of Mathematics and Computer Science (ICMC)} \\
\textit{University of São Paulo (USP)}\\
São Carlos, SP, Brazil \\
jecs89@usp.br}
}

\maketitle

\begin{abstract}
Complex Networks are a good approach to find internal relationships and represent the structure of classes in a dataset then they are used for High Level Classification.
Previous works use K-Nearest Neighbors to build each Complex Network considering all the available samples. 
This paper introduces a different creation of Complex Networks, considering only sample which belongs to each class. And metric is used to analyze the structure of Complex Networks, besides an optimization approach to improve the performance is presented. Experiments are executed considering a cross validation process, the optimization approach is performed using grid search and Genetic Algorithm, this process can improve the results between 5 and 10\%.
\end{abstract}


\section{Introduction}

Complex Networks (CN) can represent internal relationships \cite{Barabasi99, barabasi2016network, newman10, chiresaire2020new} of complex systems. During the building of CN, the process can capture existent connections between entities/objects of one class. Then, after the creation is possible to check if one element belongs or not to one class, considering a closeness metric. Therefore, it is an advantage considering traditional algorithms of Machine Learning for classification tasks, i.e. Multilayer Perceptron, Support Vector Machine, some algorithm which needs an optimization process to adjust an error function.

It is frequent in classification tasks consider that each feature, or variable can contribute to the final classification label in different proportion. Then, it is possible to assign or add a weight related to each column or variable, and this can be rewrite as an optimization problem. The objective is to get higher or lower metric through searching of these weights. This process can be solved using Evolutionary Algorithms, i.e. Genetic Algorithm (GA), because they are well-know of being useful for this kind of problems.

The process of building a Complex Network can be performed by other methods, in the state of art are: b-matching \cite{Jebara2009},  linear neighborhood \cite{Wang2008} and methods based on single linkage \cite{Cupertino2013}.

The proposal of this paper is to present a new way of building a Complex Network adding a criterion of optimization considering weight contribution of each feature. The results of this work are:
\begin{itemize}
\item New approach to build a Complex Network considering class independence
\item New metric considering shortest path in a graph
\item Improving Complex Network Building using an Optimization scheme
\end{itemize}

\section{Methods}

\begin{figure*}[!hbpt]
\centerline{
\includegraphics[width=0.95\textwidth]{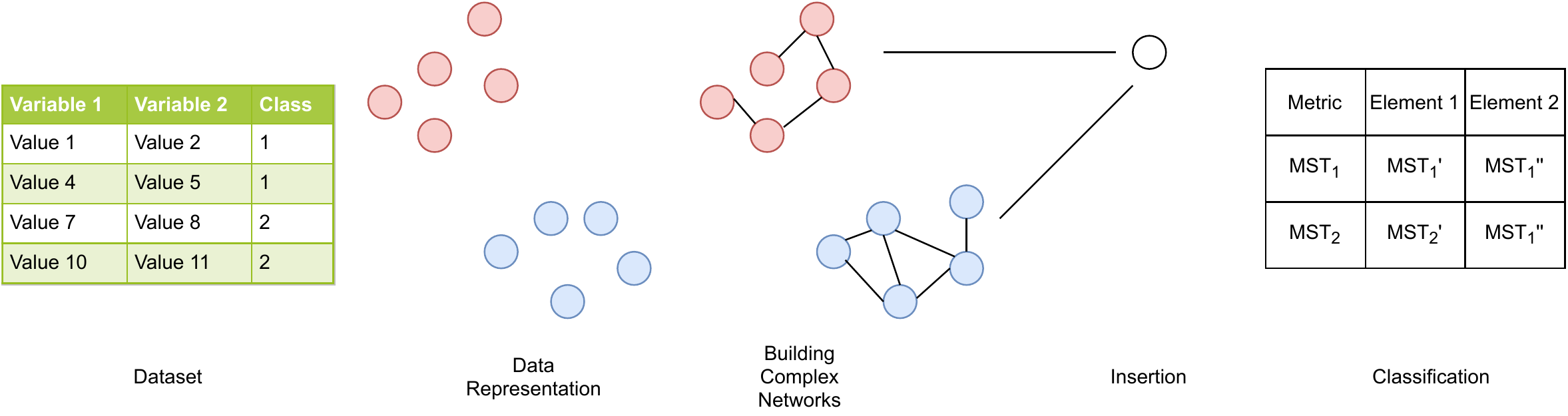}
}
\caption{Proposal methodology}
\label{fig:proposal}
\end{figure*}

The actual paper presents the building process of Complex Networks and Optimization approach to tune the structure up to improve the performance. The figure \ref{fig:proposal} summarizes these steps.
 

\subsection{Preparing dataset}

The dataset can have different scale for the variables or columns involved, then a scaling step is performed to have similar distributions. And oversampling step is used to balance dataset and have similar number of samples per class.

\subsection{Building Complex Networks}

The dataset is splitted in two parts following classic train-test process, but this proposal does not have a training process, only one building process is done.
Let separate samples per class, and create a Complex Network considering each sample as node and using a Euclidean distance to join all the nodes and assign the corespondent weight, a full connected graph.

After the creation of each Complex Network (CN), using the adjacency matrix of each CN can be executed a pruning process. This step excludes edges/connections with values higher then $\theta$*median of the weights, to tune the structure up. And after of several experiments, the proper values is $0.80$.

\subsection{Evaluation of Structure}

Previous proposal used traditional CNs metrics, i.e. neighborhood degree, asortativity and others. Following their results and discussion, is open the proposal of a More Sensitive Metric.  
Therfore, after the building of CNs are calculated the Minimum  Spanning Trees. 

An undirected graph $G = (V, E)$, where $V$ are the vertexes and $E$ are the edges which connect vertexes, i.e. $(u,v) in E$ and there are a associated cost for each edge w(u,v). Then, a path which connects all the vertexes and the total weight is minimum, with no presence of cycles, this is name minimum spanning tree (MST) \cite{cormenbook}.

\begin{equation}
    w(T) = \sum_{(u,v) \in T} w(u,v)
\end{equation}

Then, MST  represents the shortest path to join all the nodes, and the addition of the distances which compound MST is calculated as metric to represent the structure the of CN. 
It is important to highlight, this step is necessary to measure the actual structure.

\subsection{Insertion and Classification}

Let use the test set of the original dataset to know if one sample/element belongs to one class, following the next reasoning: "one element which belongs to one class, after insertion will produce less impact than one element of other class"

Then, each element is inserted to each CN and calculated the distances between all the elements of this CN and this new element. After of the insertion, a new MST' is calculated for each CN. Then, a difference is calculated between MST and MST' for each CN, and the minimum value determines to which class the element belongs.

And additional step to improve the performance is the optimization of the structure.

\subsection{Optimization of the Building}

Considering that features contribute in different levels or have different importance, the proposal includes to consider a weight for each feature, to get the level of importance or contribution following an optimization approach. The objective is to maximize the precision of the proposal.

\begin{equation}
max \quad F(x) = w_1 * feat_1 + w_2 * feat_2 + \ldots + w_n * feat_n
\end{equation}

In this proposal, Genetic Algorithm was used to find these weights.

\section{Results}

\subsection{Artificial Dataset}
\label{subsection:artdataset}

\begin{figure}[H]
  \centerline{
  \includegraphics[width = 0.5\columnwidth]{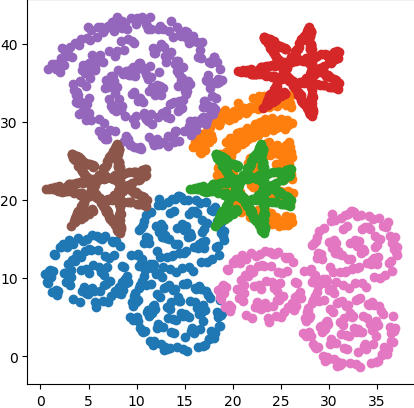}
  }
\caption{Visualization of Dataset (7 classes) and results}
\label{fig:data7}
\end{figure}

An artificial dataset is creating considering some figures, i.e. spirals, stars. Besides, an overlapping is present to present challenge to the algorithms for classification tasks, see Fig. \ref{fig:data7}

A comparison is performed with classical Machine Learning algorithm, i.e. Multilayer Perceptron, Decision Tree, Logistic Regression, Naive Bayes, Gradient and bagging, boosting.

\begin{figure}[H]
  \centerline{
  \includegraphics[width = 0.75\columnwidth]{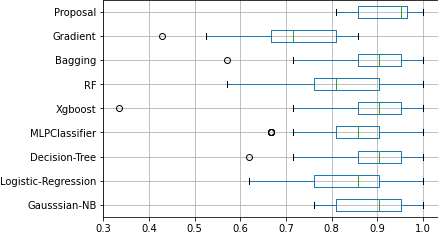}
  }
\caption{Experiment Results}
\label{fig:results7}
\end{figure}

Following the approach to optimize the building of CN, a grid search is performed. It is important to highlight the time than this process could take.

\begin{figure}[H]
  \centerline{
  \includegraphics[width = 0.75\columnwidth]{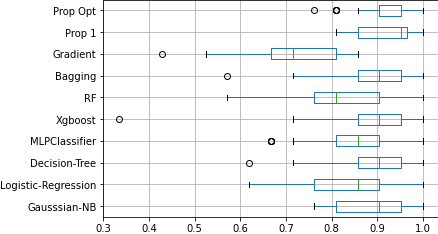}
  }
\caption{Optimization results vs previous results}
\label{fig:results7_}
\end{figure}

\subsection{Datasets}
\label{subsection:datasets}

The chosen dataset are available in \cite{romano2021pmlb}, after a search were selected dataset with at least three hundreds samples per class, numerical data and at least two classes. Therefore, were selected: magic, satimage, sleep and phoneme. Then, a brief description of the datasets is present in Tab. \ref{datasets}. The website presents a filter bars to select the datasets, but it was necessary to go through the sites and description of datasets to find the proper for the experiments.


\begin{table}[!h]
\centering
\begin{tabular}{|c|c|c|c|}
\hline
\multicolumn{4}{|c|}{\textbf{Dataset PMLB}}                                                                                                         \\ \hline
\textbf{Name} & \textbf{Samples Class}                                                                      & \textbf{Variables} & \textbf{Classes} \\ \hline
magic         & 0: 12332, 1: 6688                                                                           & 10                 & 2                \\ \hline
satimage      & \begin{tabular}[c]{@{}c@{}}1: 1533, 2: 703, 3: 1358,\\  4: 626, 5 707, 7: 1508\end{tabular} & 36                 & 7                \\ \hline
sleep         & \begin{tabular}[c]{@{}c@{}}0: 21359,1: 9052,2: 52698,\\ 3: 10832,5: 11967\end{tabular}      & 13                 & 5                \\ \hline
phoneme       & 0: 3818, 1: 586                                                                             & 5                  & 2                \\ \hline
\end{tabular}     
\caption{Dataset Description}
\label{datasets}
\end{table}

\begin{figure}[hbpt]
  \centerline{
  \includegraphics[width = 1.0\columnwidth]{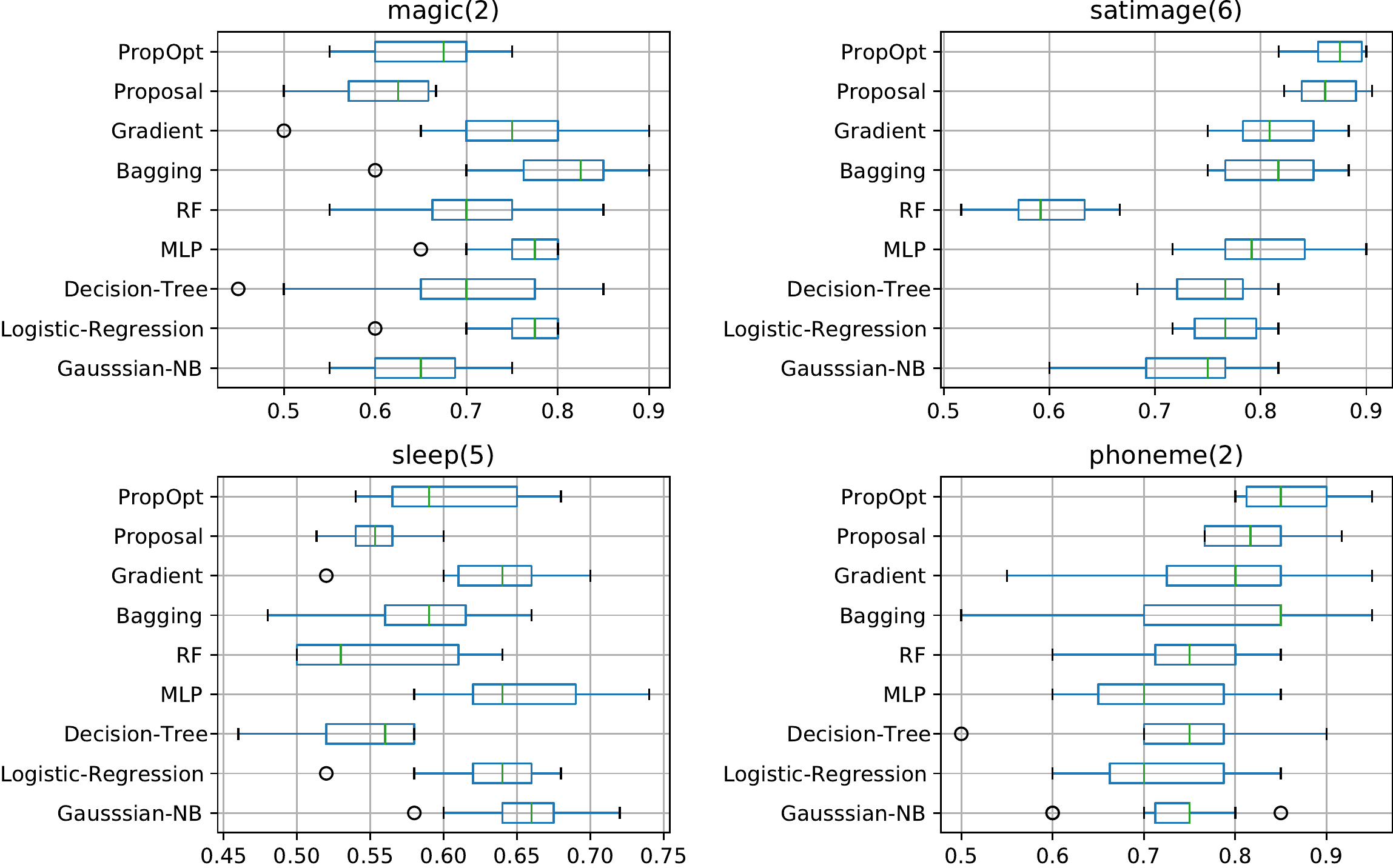}
  }
\caption{}
\label{fig:results_}
\end{figure}

After the experiments using a cross validation of k=10, the results are presented in Fig \ref{fig:results_}. The proposal without optimization is named $Proposal$ and with optimization is $PropOpt$.

\subsection{Analysis of the results}

Considering the results of section \ref{subsection:artdataset}, the proposal outperform the other algorithms is only check the minimum values. Besides, the median is higher and the limits of the boxplot are better. These results show the proposal can have good performance. After adding optimization approach, the proposal reduces the limits of the boxplot, and better performance than all the algorithms. This is related to the contribution of each variable to the classification task.

By the other hand, results of the subsection \ref{subsection:datasets} show the proposal got the best values with satimage, phoneme datasets. Besides, the optimization approach can improve the results up to 10\%.
But, with datasets: magic and sleep is in the last three last positions. In spite of these results, the optimization approach can improve them up to 10\%. 

\section{Conclusion}

In conclusion, the proposal to separate and create independent CN for each class can present good results, the experiments can support this affirmation. The metric introduced to MST to calculate the shortest path in the CN helps the global performance. Finally, the optimization approach can optimize the results up to 10\% to the original proposal. 



\bibliographystyle{IEEEtran}
\bibliography{biblio.bib}


\end{document}